\documentclass[final,3p,times,twocolumn]{elsarticle}
\usepackage{graphicx}
\usepackage{url}
\usepackage{slashbox}
\usepackage{wrapfig}

\usepackage{amssymb}
\journal{Journal of Robotics and Autonomous Systems (Elsevier)}

\begin{document}

\begin{frontmatter}



\title{Extracting Semantic Indoor Maps from Occupancy Grids}
%%(Semantic Exploration of Indoor Maps)

\author{Ziyuan Liu\fnref{label2}}
\address{Institute of Automatic Control Engineering, Technische Universit\"at M\"unchen, Munich, Germany\\ Institute for Advanced Study, Techniche Universit\"at M\"unchen, Lichtenbergstrasse 2a, D-85748 Garching, Germany}
\ead{ziyuan.liu@tum.de}
%%\address{Institute of Automatic Control Engineering, Technische Universit\"at M\"unchen, Germany}
\fntext[label2]{Corresponding author. Postal address: Karlstr. 45, room 3015, 80333, Munich, Germany. Telephone: +49-89-289-26900. Fax: +49-89-289-26913.}

\author{Georg von Wichert}
\ead{georg.wichert@siemens.com}
\address{Siemens AG, Corporate Research \& Technologies, Munich, Germany \\ Institute for Advanced Study, Techniche Universit\"at M\"unchen, Lichtenbergstrasse 2a, D-85748 Garching, Germany}

\begin{abstract}

The primary challenge for any autonomous system operating in realistic, rather unconstrained scenarios is to manage the complexity and uncertainty of the real world. While it is unclear how exactly humans and other higher animals master these problems, it seems evident, that abstraction plays an important role. The use of abstract concepts allows to define the system behavior on higher levels. %This abstractly defined behavior would be applicable to a wide range of situations and thus increase the overall robustness of the system. 
%In addition, we directly address the uncertainty related issues by strictly following a probabilistic approach. 
In this paper we focus on the semantic mapping of indoor environments. We propose a method to extract an abstracted floor plan from typical grid maps using Bayesian reasoning. The result of this procedure is a probabilistic generative model of the environment defined over abstract concepts. It is well suited for higher-level reasoning and communication purposes. We demonstrate the effectiveness of the approach using real-world data.
\end{abstract}


\end{frontmatter}

% \linenumbers

\section{Introduction}
\label{int}

The primary challenge for any autonomous system operating in realistic, rather unconstrained scenarios is to manage the complexity and uncertainty of the real world. In robotics this holds, as soon as the robots leave the carefully engineered production environments in which they have been so successful in the past decades.

The typically high degree of uncertainty in real-world environments, that makes a robot's life so hard, comes from the following sources: the limited measurement accuracy and other limitations of the system's sensors, modeling errors and purposefully made simplifications in the system's internal representations, unobserved environment dynamics and random effects in action execution. While it is unclear how exactly humans and other higher animals master these problems, it seems evident, that abstraction plays an important role. The use of abstract concepts allows to define the system behavior on higher levels and independently of the exact setting of the environment and the exact sensor readings. %This abstractly defined behavior would be applicable to a wider range of situations and thus increase the overall robustness of the system.

In this study we address the first two of the problems mentioned above, in that we provide the system with a limited capability of abstraction allowing for a higher-level understanding of its environment. In addition, we directly address the uncertainty related issues by strictly following a probabilistic approach that explicitly models and keeps track of the uncertainty associated with any variables of the problem.

As a by-product, the system's capability to use predefined concepts will ease cooperation in mixed human-robot tasks, since a common language used by both the human and the robot is a precondition for efficient exchange of information between both parties. This is however not addressed in this paper.

To illustrate the general idea, we use an example from an indoor navigation scenario, namely the semantic analysis of the commonly used occupancy grid maps. The objective of the presented method is to provide an abstracted, semantically annotated but still probabilistic map of the indoor environment. For this purpose, we first use a robot -- equipped with a 2D laser scanner -- to build an occupancy grid of the environment using a standard SLAM method \cite{grisetti2007improved} and then employ the procedure described in the reminder of this document to extract the semantic information. To do this, we use a Markov Chain Monte Carlo (MCMC) based sampling technique \cite{zhu2000integrating} to generate samples from the probability density function capturing the distribution of probable worlds the robot could encounter. The maximum posterior solution could then be used as an estimate of what the world semantically looks like.

\section{Problem Formulation}
\label{pro}

\begin{figure}
	\centering
	\includegraphics[width=0.9\columnwidth]{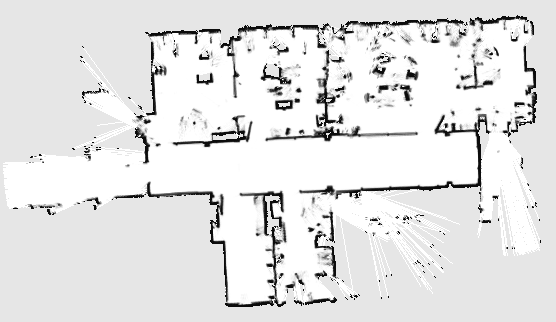}
    \caption{A typical occupancy grid map of an indoor environment, obtained from the Robotics Data Set
    Repository (Radish)~\cite{Radish}.}
    \label{figure:gridmap1}
\end{figure}

Most of todays' mapping approaches aim to construct a globally consistent, metric map of the robot's operating environments. See Fig.~\ref{figure:gridmap1} for a typical result. Such maps enable the robot to localize itself with respect to the environment and thus determine its global pose in an assumed flat world with an accuracy of typically a few centimeters in translation and below one degree in rotation. Based on this capability, the robot can also plan a path and navigate towards a goal, that will also be specified by its metric position in the global map reference frame. However, the robot does not understand its environment in terms of typical semantic concepts like rooms, corridors or functionally enriched concepts like a kitchen or living room. Furthermore, the robot does not understand relations like adjacency, connectivity via doors, or properties like rectangularity that -- if known to be relevant to the given environment -- could help to build the maps in the first place.

Our work aims at extracting such semantic models of the environment from the more or less raw sensor data. In the context of this paper, we assume, that a map, like the one depicted in Fig.~\ref{figure:gridmap1}, was already constructed using one of the proven methods available for this purpose~\cite{grisetti2007improved}.

Assigning semantics to spatial maps in robotics has not been looked at as intensely as the metric or topological mapping. Still, several important contributions to the field have already been made. They can be clustered into two major groups. The first group consists of methods based on place labeling, some notable examples are \cite{wolf2008semantic, pronobis2010cogsys, friedman2007voronoi,goerke2009building,mozos2006supervised,mozoz2007supervised,Viswanathan2009,Granda2010}. These methods assign semantic labels to places or regions of the accessible work space of the robot. They are very much in the tradition of \cite{thrun1996integrating} or \cite{kurz1996constructing}.

A second group is formed by approaches assigning semantic labels to parts or objects of the perceived structure of the environment, like traversable terrain, trees or similar structures in outdoor environments or walls, ceilings, and doors in indoor settings~\cite{limketkai2005relational, douillard2008laser, nuechter2008towards,tong2010,wang2011,krishnan2010,persson2007,Jebari2011,case2011}. 

In addition to the two groups mentioned above, there are also other approaches. The approach of \cite{ranganathan2007semantic} semantically models places via objects. In \cite{laviers2004}, a method is proposed, which explores the environment in a room-by-room style and fits the explored map part into polygons. Tapus and Siegwart \cite{tapus2005} build a map of the environment based on so called fingerprints of explored places. Lim et. al. \cite{Lim2011} introduce an ontology-based method that integrates low-level data with high-level constraints to represent the knowledge as a semantic network. 

\begin{figure*}[!htb]
\centering
\includegraphics[height=11cm]{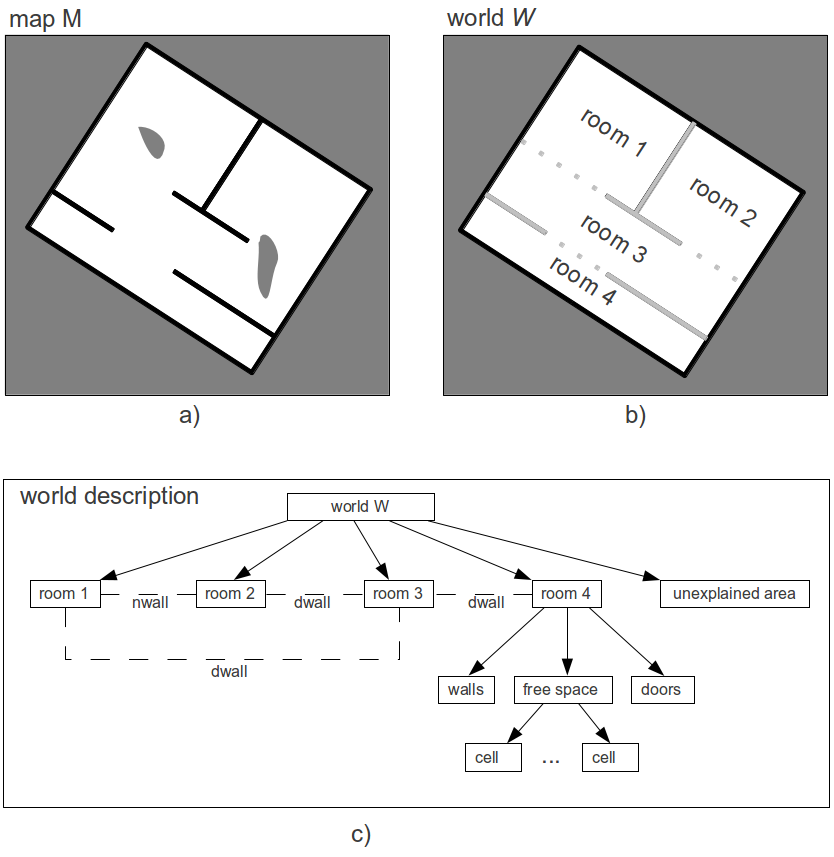}
\caption{a) A simplified occupancy grid map: Unexplained area is drawn in grey, free space is drawn in white. Occupied area is drawn in black. b) A possible floor plan represented as a scene graph ($W$): The world is divided into four rooms and the corresponding unexplained area. The connectivity is given by the wall types (dwall: a wall that has one or more doors on it; nwall: a wall that separates two rooms but does not contain a door on it; bwall: a wall that just serves as boundary). A partially dotted line in light-gray indicates a dwall, where the dotted part is the door, and the solid line part is the rest of the wall. A light-gray line (without dots) shows one nwall, and black stands for a bwall. c) The semantic description of the world in form of the scene graph: Directed links connect nodes. The dashed lines represent connectivity. Like room 4, each room has three child nodes: walls, free space, and doors. Note that the lowest level of node in the tree structure is the grid cell that belongs to walls, free space and doors.}
\label{figure:example}
\end{figure*}

Different from those methods mentioned above, we aim to construct a probabilistic generative model of the world around the robot, that is essentially based on abstract semantic concepts but at the same time allows to predict the continuous percepts that the robot obtains via its noisy sensors. This abstract model has a form similar to a scene graph, a structure which is widely used in computer graphics. The scene graph (see Fig.~\ref{figure:example} c) in our case consists of rooms and doorways connecting the rooms and can be visualized as a classical floor plan(see Fig.~\ref{figure:example} b).

The scene graph and thus also the semantically annotated world state is denoted by a vector of hidden parameters $W$ specifying the world state, that generated the occupancy map $M$ we are currently looking at. In the Bayesian framework we can use a maximum posterior approach to infer the most probable state $W^*\in\Omega$ from the space of probable worlds $\Omega$ given the map $M$. 

\begin{equation}
W^*=\arg\!\max_{\!\!\!\!\!\!\!\!\!\!\!_{W\in\Omega}} \, p(W|M),
\label{eq:argmax}
\end{equation}

with 

\begin{equation}
p(W|M)\propto p(M|W)p(W).
\label{eq:generative}
\end{equation}

Here $p(W|M)$ is the posterior distribution of $W$ given the known map $M$ and $p(W)$ is the prior specifying, which worlds $W$ are possible at all. $p(M|W)$ is the likelihood function describing how probable the observed map $M$ is, given the different probable worlds represented by a parameter vector $W$. The actual semantic model is represented in the structure of the parameter vector $W$, while semantically relevant constraints go into the prior $p(W)$. 

\section{A Generative Model for Occupancy Grids}
\label{gen}

In our case $W$ contains the scene graph, i.e. the parameters of a floor plan: number of rooms, their dimensions and connectivity, the location of doorways. Here, rooms are defined as rectangular space that is enclosed by four \emph{walls}, and walls are line segments defined by two end points. Connectivity indicates the spatial relationship of different rooms, and it is expressed in the types of walls: \emph{wall with door} (dwall), \emph{neighbor wall} (nwall) and \emph{boundary wall} (bwall). A dwall is a wall that has one or more doors on it. The term door acutally refers to a door opening in a wall, which is a line segment made up by free space that can be passed so as to enter another room.  A nwall separates two rooms but does not contain a door on it, whereas a bwall is a wall that just serves as the boundary of the world.

Certain a-priori assumptions about some properties of the structured world are made based on context knowledge as follows:

\begin{enumerate}
	\item[1)] a room has four walls and possesses a rectangular shape.
	\item[2)] a room has at least one door, and a door is placed on a wall.
	\item[3)] each cell in the map should only belong to one room.
\end{enumerate}

These a-priori constraints are enforced by means of the prior $p(W)$ in our generative model (\ref{eq:generative}).
The prior penalizes worlds that are not fully compliant with the above assumptions:

\begin{equation}
p(W)=\alpha_1\times\alpha_2\times\alpha_3,
\end{equation}
where $\alpha_1,\alpha_2$ and $\alpha_3$ are the corresponding penalization terms for the point 1), 2) and 3) of the prior information respectively, and they are defined as follows:

\begin{equation}
\alpha_1=\left\{\begin{array}{lc}
\psi_1^\theta,\textrm{conflict with point 1)},\\
1,\textrm{otherwise},\\
\end{array}
\right.
\end{equation}

\begin{equation}
\alpha_2=\left\{\begin{array}{lc}
\psi_2,\textrm{conflict with point 2)},\\
1,\textrm{otherwise},\\
\end{array}
\right.
\end{equation}

\begin{eqnarray}
\alpha_3&=&\prod\limits_{c(x,y)\in M}\psi_3^{\gamma(c(x,y))},\nonumber\\
\gamma(c(x,y))&=&\left\{\begin{array}{lc}
\sigma(c(x,y))-1,\sigma(c(x,y))>1,\\
0,\textrm{otherwise},\\
\end{array}
\right.
\end{eqnarray}
where $\psi_1,\psi_2$ and $\psi_3$ are penalization terms with $\psi_1,\psi_2,\psi_3\in(0,1)$. $\theta$ is the number of pairs of adjacent walls whose included angle is not 90 degree ($\pm$tolerance). $c(x,y)$ denotes one grid cell in the map $M$. $\sigma(c(x,y))$ indicates the number of rooms, to which $c(x,y)$ belongs. $\alpha_3$ is a cell-wise penalization of the overlap between different rooms, i.e. if there is no overlap in one cell $c(x,y)$, then $\sigma(c(x,y))$ is equal to 0 or 1, in which case $\gamma(c(x,y))$ is 0 (no penalization in cell $c(x,y)$). Otherwise, if $\sigma(c(x,y))$ is bigger than 1, which means the cell $c(x,y)$ belongs to more than one room, then $\gamma(c(x,y))$ is bigger than 0 (penalization in cell $c(x,y)$). For the experiments shown in Section \ref{exp}, the penalization terms are set as follows:$\psi_1=0.9,\psi_2=0.9,\psi_3=0.6$.

For our generative model, we need to specify the likelihood function $p(M|W)$ additionally. Since $M$ is represented by an occupancy grid with statistically independent grid cells $c \in M$, we only need to come up with a model $p(c|W)$ for all cells at their locations $(x,y)$ in the map M:

\begin{equation}
p(M|W) = \prod_{c(x,y) \in M} p(c(x,y)|W).
\label{eq:prod}
\end{equation}

For our model $p(c(x,y)|W)$, we first discretize the cell state $M(x,y)$ by classifying the occupancy values into three classes 
``occupied=2", ``unexplained=1" and ``free=0" so as to generate the classified map $C_M(x,y)$ according to:

\begin{equation}\label{equ:classify}
C_M(x,y)=\left\{\begin{array}{lcc}
2,\quad 0\leq M(x,y)\leq h_o,\\
1, \quad h_o<M(x,y)\leq h_u,\\
0,\quad h_u<M(x,y),
\end{array}
\right.
\end{equation}
where $h_o$ and $h_u$ are the intensity thresholds for occupied and unexplained grid cells. Based on our world model $W$ we can also predict expected cell states $C_W(x,y)$ accordingly:

\begin{equation}
C_W(x,y)=\left\{\begin{array}{lcc}
2,\quad (x,y)\in S_w,\\
1,\quad (x,y)\in S_u,\\
0,\quad (x,y)\in S_f,
\end{array}
\right.
\end{equation}
where $S_w, S_u$ and $S_f$ are the set of all the wall cells, unknown cells and free space cells in the world $W$ respectively. $p(c(x,y)|W)$ can then be represented in the form of a lookup-table.

\begin{table*}[htb]
	\centering
	\begin{tabular}{|c||*{3}{c|}}\hline
	\backslashbox{$C_W(x,y)$}{$C_M(x,y)$}
	&\makebox[6em]{0 (occupied)}&\makebox[6em]{1 (unexplained)}&\makebox[6em]{2 (free)}\\\hline\hline
	%%&0&1&2\\\hline\hline
	0 (wall)&0.8&0.1&0.1\\\hline
	1 (unknown)&0.1&0.8&0.1\\\hline
	2 (free)&0.1&0.1&0.8\\\hline
	\end{tabular}
	\caption{The lookup table for $p(c(x,y)|W)$.}
	\label{TAB:mapping table}
\end{table*}

In principle the likelihood $p(c(x,y)|W)$ plays the role of a sensor model. In our case it captures the quality of the original mapping algorithm producing the grid map (including the sensor models for the sensors used during the SLAM process), and could be learned from labeled training data. However, for the experiments described in section~\ref{exp} we used the empirically determined values given in Table \ref{TAB:mapping table}.

\section{Searching The Solution Space}
\label{DDMCMC}

For solving equation~(\ref{eq:argmax}) we need to efficiently search the large and complexly structured solution space $\Omega$. Here we adopt the approach of \cite{zhu2000integrating}, who propose a data driven Markov chain Monte Carlo (MCMC) technique for this purpose. The basic idea is to construct a Markov Chain that generates samples $W_i$ from the solution space $\Omega$ according to the distribution $p(W|M)$ after some initial burn-in time. One popular approach to construct such a Markov chain is the Metropolis-Hastings (MH) algorithm \cite{bishop2007machine,Siddhartha}. In MCMC techniques the Markov chain is constructed by sequentially executing state transitions (in our case from a given world state $W$ to another state $W'$) according to a transition distribution $\Phi(W'|W)$ of the sub-kernels. An example of $\Phi(W'|W)$ is given in Table \ref{TAB:Transition}. In order for the chain to converge to a given distribution, it has to be reversible and ergodic~\cite{bishop2007machine}. The MH algorithm achieves this by generating new samples in three steps. First a transition is proposed according to $\Phi(W'|W)$, subsequently a new sample $W'$ is generated by a proposal distribution $Q(W'|W)$, and then it is accepted with the following probability:

\begin{equation}
\lambda(W,W') = \min\left( 1, \frac{p(W'|M) Q(W|W')}{p(W|M) Q(W'|W)} \right)
\label{eq:MH}
\end{equation}

The resulting Markov chain can be shown to converge to $p(W|M)$. However the selection of the proposal distribution is crucial for the convergence rate. Here, we follow the approach of \cite{zhu2000integrating} to propose state transitions for the Markov chain using discriminative methods for the bottom-up detection of relevant environmental features (e.g. walls, doorways) and constructing the proposals based on these detection results. \cite{zhu2000integrating} created the term data driven MCMC for this procedure.

\subsection{MCMC Kernels}

In order to design the Markov chain in form of the Metropolis-Hastings algorithm, the kernels that modify the structure of the world are arranged as reversible pairs. Currently, we use four pairs of kernels, and these include:

\begin{itemize}
	\item Kernel pair 1: ADD or REMOVE one room.
		\begin{itemize}
			\item ADD: draw one new room from certain candidates, then try to add this room to the world.
			\item REMOVE: try to cancel one existing room from the world.
		\end{itemize}
	\item Kernel pair 2: SPLIT one room or MERGE two rooms.
		\begin{itemize}
			\item SPLIT: try to decompose one existing room into two rooms.
			\item MERGE: try to combine two existing rooms, and generate one new room out of them.
		\end{itemize}
	\item Kernel pair 3: SHRINK or DILATE one room.
		\begin{itemize}
			\item SHRINK: try to move one wall of one room along certain orientation, so that the room becomes smaller.
			\item DILATE: similarly to SHRINK, move one wall of one room, so that the room becomes bigger.
		\end{itemize}
	\item Kernel pair 4: ALLOCATE or DELETE one door
		\begin{itemize}
			\item ALLOCATE: draw one door from the door candidates that are provided by door detector, then try to assign it to two existing rooms.
			\item DELETE: cancel one assigned door.
		\end{itemize}
\end{itemize}

\begin{figure*}[htb]
	\centering
  	\includegraphics[height=13cm]{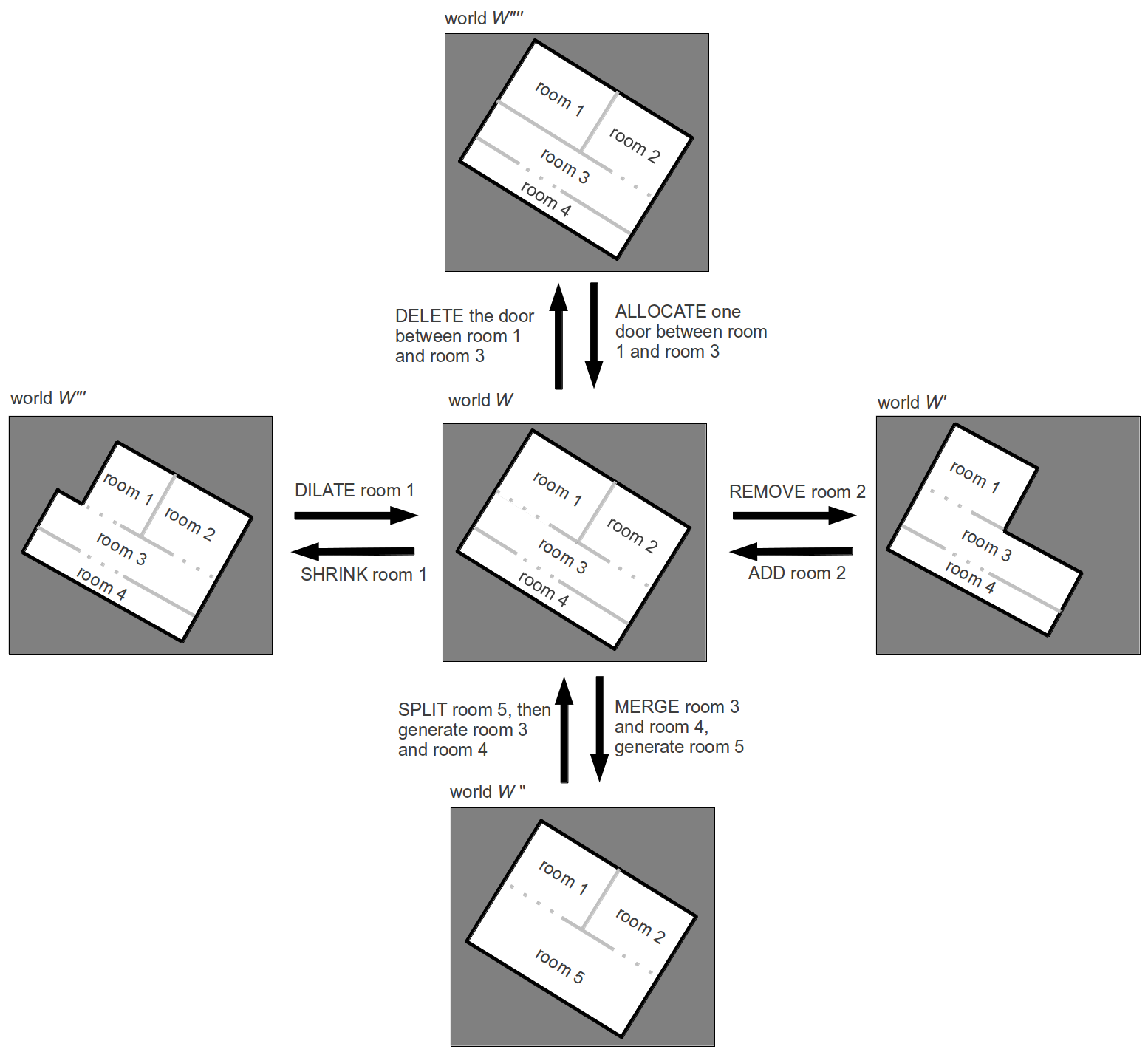}
    \caption{Reversible MCMC kernel pairs: ADD/REMOVE, SPLIT/MERGE, SHRINK/DILATE and ALLOCATE/DELETE.}
   	\label{figure:mcmc-kernels}
\end{figure*}

Fig. \ref{figure:mcmc-kernels} shows an example of the four reversible MCMC kernel pairs, using the same simplified occupancy grid map as in Fig. \ref{figure:example}. The world $W$ can transit to $W^{'}$, $W^{''}$, $W^{'''}$ and $W^{''''}$ by applying the sub-kernel REMOVE, MERGE, SHRINK and DELETE, respectively. By contrast, the world $W^{'}$, $W^{''}$, $W^{'''}$ and $W^{''''}$ can also transit back to $W$ using corresponding reverse sub-kernels.

\subsection{Discriminative Generation of Chain Transition Proposals}

As mentioned above, we use discriminative methods to propose candidates for MCMC kernels, so as to accelerate the convergence of the Markov chain. In the following, we define and explain the  discriminative methods used with respect to corresponding MCMC kernels.

\subsubsection{ADD}

Currently, two room detectors are adopted to propose room candidates for ADD: a \emph{wall based room} (WBR) detector and a \emph{free space room} (FSR) detector. In the context of topological navigation several approached for the online detection of rooms have been proposed~\cite{buschka2002virtual,wichert99}. Here, we detect the rooms offline based on the input map.

\paragraph{WBR detector}

\begin{figure*}[!htb]
	\centering
  	\includegraphics[width=0.8\textwidth]{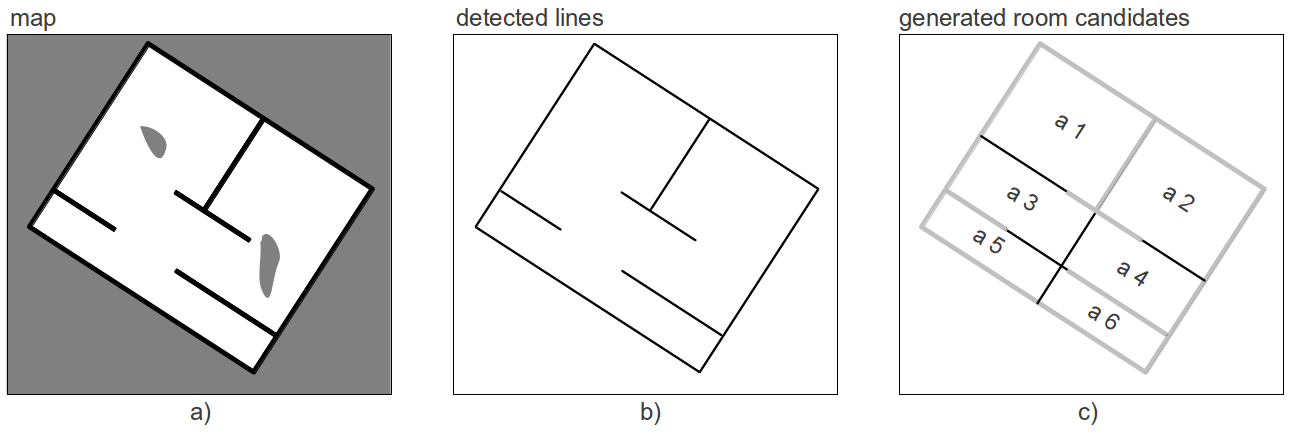}
    \caption{The WBR detector. a) A simplified occupancy grid map. b) The line segments detected by the Hough line detection. c) Divide the map into sub-areas a1, a2, a3, a4, a5 and a6, by extending the detected lines. Generate room candidates out of these sub-areas. The detected line segments are shown in light-gray.}
   	\label{figure:ADD}
\end{figure*}

This room detector makes use of the well known Hough Transform \cite{hough1962} for edge based wall detection and generates room candidates using the detected line segments according to the following procedure: We extend the detected line segments so as to divide the map into sub-areas, then these sub-areas are recombined to give new rooms which obey our prior information. One example of the WBR room detector is demonstrated in Fig. \ref{figure:ADD}. Here, the map is divided into six sub-areas: a1, a2, a3, a4, a5 and a6, and these sub-areas are recombined to generate 18 room candidates, which are: a1, a2, a3, a4, a5, a6, a1a2, a3a4, a5a6, a1a3, a3a5, a2a4, a4a6, a1a3a5, a2a4a6, a1a2a3a4, a3a4a5a6 and a1a2a3a4a5a6. We sample from the set of all candidate rooms in a resampling style \cite{kitagawa}. Each of the generated room candidates are weighted according to how well their walls match the observations provided by the occupancy grid map. The weight of a room $\omega_r$ is defined as the lowest wall weight $\omega_{w_{j}}$ among its four walls, where $j,j\in\{r_1,r_2,r_3,r_4\}$, indexes the wall, with $r_i,i\in\{1,2,3,4\}$, indicating the $i$th wall of room $r$:

\begin{equation}
\omega_r=\min_{j\in\{r_1,r_2,r_3,r_4\}}{\omega_{w_{j}}}.\label{equ:roomweight}
\end{equation}

The wall weight $\omega_{w_{ j}}$is calculated as:

\begin{equation}
\omega_{w_{ j}}=\frac{n(w_{ j})}{l(w_{ j})},\label{equ:wallweight}
\end{equation}
where $l(w_{ j})$ indicates the length of wall $w_{ j}$ and can be computed from the coordinates of its two end points $(x_{w_{ j,1}},y_{w_{ j,1}}),(x_{w_{ j,2}},y_{w_{ j,2}})$:

\begin{equation}
l(w_{ j})=\sqrt{(x_{w_{ j,1}}-x_{w_{ j,2}})^2+(y_{w_{ j,1}}-y_{w_{ j,2}})^2}.\label{equ:length}
\end{equation}

The term $n(w_{ j})$ counts the number of wall cells that match with the map:

\begin{equation}
n(w_{ j})=\sum\limits_{(x,y)\in w_{ j}}t(x,y),\label{equ:number}
\end{equation}

where

\begin{equation}\label{equ:01}
t(x,y)=\left\{\begin{array}{lc}
1,\quad C_{M}(x,y)=0,\\
0,\quad \textrm{otherwise}.
\end{array}
\right.
\end{equation}

Having obtained the weights of all the room candidates, we implement $Q(W'|W)$ by sampling from the candidates according to their cumulative weights $A_r$. First, we normalize their weights $\omega_r$:

\begin{equation}
\omega^{'}_r=\frac{\omega_r}{\sum\limits_{r\in B}\omega_r},\label{equ:normalize}
\end{equation}
where $B$ indicates the set of all the room candidates generated by the WBR detector. Then, we calculate the cumulative weights $A_r$ for room $r$:

\begin{equation}
A_r=\sum\limits_{i=1}^r\omega_i.
\end{equation}

Finally, we can draw a room candidate $n$ out of $B$, by generating a random number $k,k\in[0,1)$,

\begin{equation}
n=\min\{i|k\leq A_i\}.\label{equ:sampling}
\end{equation}

\paragraph{FSR detector}

\begin{figure*}[!htb]
	\centering
  	\includegraphics[width=0.8\textwidth]{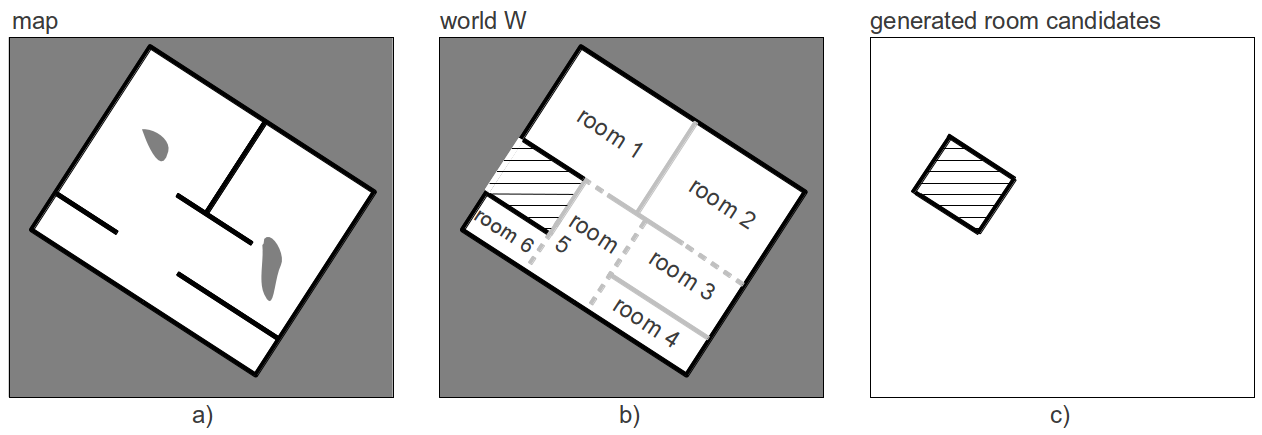}
    \caption{FSR room detector. a) A simplified occupancy grid map. b) The current world $W$: the shaded area is not explained. c) $C_M''$: cancel the already explained regions from $C_M'$ and detect unexplained regions based on connected-component analysis, then generate room candidates out of these detected regions. In this example, a room candidate is generated from the shaded area.}
   	\label{figure:ADD2}
\end{figure*}

Sometimes rooms will be missed by the edge-based procedure mentioned above. This is often the case for rooms only partially explored during grid map construction or when walls have been obstructed by furniture and thus not have been perceived by the laser scanner. We therefore use an alternative method for room detection. This detector works on the basis of connected-components analysis and is referred to as free space room (FSR) detector. If there are still regions of the map which are not explained by the world after many MCMC steps (4000 steps in the experiments), then we try to find them using the FSR detector as follows: 1) make a copy of the classified map, which we denote as $C_M'$; 2) cancel the regions that are already explained by the current semantic world $W$ from $C_M'$, we denote the rest of $C_M'$ as $C_M''$; 3) detect unexplained regions in $C_M''$ based on connected-components analysis \cite{chang04}, and generate room candidates out of them. Fig. \ref{figure:ADD2} shows one example of the FSR detector. Here, the world $W$ does not cover the shaded area in the map, then we detect it using the FSR detector and generate room candidates out of it. Having generated the room candidates, we use the same sampling technique as done with the WBR detector to propose room candidates ((\ref{equ:roomweight}) to (\ref{equ:sampling})). The number of MCMC steps after which the FSR detector is activated should be big enough so that each of the unexplained regions is relatively small and can be easily used to generate new room candidates. 

%The proposal probability $Q_1(W^{'}|W)$ for ADD is equal to the normalized weight $\omega^{'}_n$ of the selected room candidate $n$:
%\begin{equation}
%Q_1(W^{'}|W)=\omega^{'}_n.
%\end{equation}
%
%For the calculation of the back proposal probability $Q_1(W|W^{'})$ of ADD, we observe the room $n$ as one member room of the world $W^{'}$ and normalize the weights of the 
%The WBR detector and the FSR detector have different functionalities in the Markov chain. The WBR detector is used as the starting point of the MCMC dynamics, and it provides candidates for the relatively unknown world, whereas the FSR detector is adopted after the Markov chain has reached a somehow steady state, so as to find the areas that cannot be proposed by the WBR detector or by other MCMC kernels.

\subsubsection{SPLIT}

For SPLIT we again use the Hough transformation based line detection to propose splitting options. Hough line detection is applied within rooms that already exist in the world $W$ (member rooms of $W$). First, a room $r$ is randomly chosen according to a uniform distribution, which means every room contained in the current world has the same chance to be chosen. Then, the Hough line detector is applied within the room $r$ to detect possible room splits. Let $E_r$ denote the set of the detected line segments within room $r$. Each detected line segment $e,e\in E_r$ is weighted, using its length $l(e)$:

\begin{equation}
\omega_e=l(e),
\end{equation}
where the length $l(e)$ is similarly calculated as done in (\ref{equ:length}). Then we normalize the weights and build the cumulative distribution of $E_r$. Furthermore, we draw one line segment out of $E_r$, as done with the WBR detector ((\ref{equ:normalize}) to (\ref{equ:sampling})).

\begin{figure*}[htb]
	\centering
  	\includegraphics[width=0.8\textwidth]{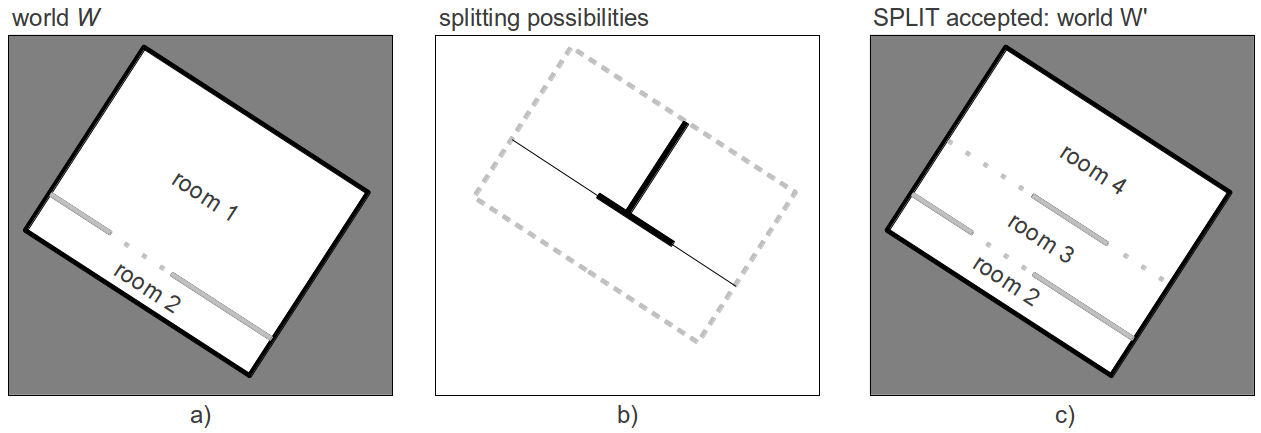}
    \caption{SPLIT sub-kernel. a) The current world $W$. b) The room 1 (light-gray dashed rectangle) is chosen for SPLIT. Two line segments are detected (black thick lines) as splitting possibilities. One of them is selected and extended to split the room (the black thin line). c) After the accepted SPLIT action, the new world $W^{'}$ is created.}
   	\label{figure:split}
\end{figure*}

Once a line segment is chosen, it is extended, so that it intersects with the walls of the room. Currently, only the case is accepted that two opposite walls of the room are intersected. The case that two neighbor walls are intersected is neglected. With the extended line segment, we propose to split the room into two rooms. The MH algorithm then decides  whether this action is accepted. A typical example of the SPLIT sub-kernel is shown in Fig. \ref{figure:split}.

\subsubsection{MERGE}

The sub-kernel MERGE is the inverse of SPLIT. It tries to combine two member rooms of the current world $W$, so as to generate a new room, then the MH algorithm decides whether the proposed new room is accepted. To do this, the first room $r$ is drawn from the set of all member rooms $R_W$ of world $W$ according to a uniform distribution, which means that each member room has the same possibility to be chosen. Additionally, a second room $s$ needs to be selected from the rest of the member rooms, $s\in R_W\setminus r$. For sampling $s$, we define a new weight $a_r(s)$, which is the reciprocal of the distance $d(c_r,c_s)$ between the center point $c_r$ of room $r$ and the center point $c_s$ of room $s$:

\begin{equation}
d(c_r,c_s)=\sqrt{(c_r.x-c_s.x)^2+(c_r.y-c_s.y)^2},
\end{equation}
where $(c_r.x,c_r.y)$ and $(c_s.x,c_s.y)$ are the grid cell index of the two center points. The weight $a_r(s)$ is calculated as follows:

\begin{equation}
a_r(s)=\frac{1}{d(c_r,c_s)}.
\end{equation}

\begin{figure*}[htb]
	\centering
  	\includegraphics[width=0.8\textwidth]{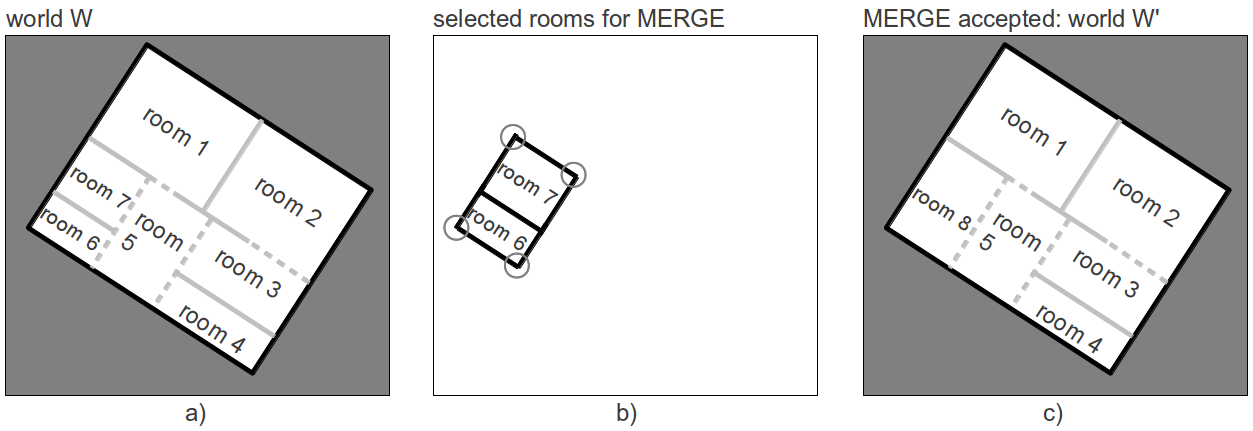}
    \caption{MERGE sub-kernel. a) The current world $W$. b) The room 6 is selected as the first room, the room 7 is selected as the second room. Here, room 7 and room 5 have better chance to be chosen as the second room than room 1, 2, 3 and 4, because room 5 and 7 are closer to room 6. The four vertices of the new room are surrounded by gray circles. c) After the accepted MERGE action, the world $W^{'}$ is generated.}
   	\label{figure:merge}
\end{figure*}

Subsequently, we normalize the weights $a_r(s)$, calculate the cumulative probability and draw the second room, as done in (\ref{equ:normalize}) to (\ref{equ:sampling}). Once the two rooms are obtained, we try to combine them into one room. The underlying idea for using $a_r(s)$ in the sampling is that the closer two rooms are spatially located, the more likely they can be combined. Fig. \ref{figure:merge} illuminates an example of MERGE.

\subsubsection{SHRINK and DILATE}

The kernel pair SHRINK/DILATE tries to move a wall $w_j$ of a member room $r$ in the current world $W$, so that this room can better match the map. Here, $j,j\in\{r_1,r_2,r_3,r_4\}$, indexes the wall, with $r_i,i\in\{1,2,3,4\}$, indicating the $i$th wall of room $r$. For selecting the room $r$ from the set of all member rooms $R_W$, we define a new weight $b_r$:

\begin{equation}\label{equ: new room weight}
b_r=\left\{\begin{array}{cl}
 \frac{1}{\omega_r}, &\frac{1}{\omega_r} \leq h_b\\
h_b, &\textrm{otherwise},
\end{array}
\right.
\end{equation}
where $\omega_r$ is the room weight defined in (\ref{equ:roomweight}). $h_b$ is a predefined threshold for the weight. Using $b_r$, a room is drawn according to (\ref{equ:normalize}) to (\ref{equ:sampling}). The reason for this is that the worse a room matches the map, the more likely it should be changed by SHRINK/DILATE. 

\begin{figure*}[htb]
	\centering
  	\includegraphics[width=0.8\textwidth]{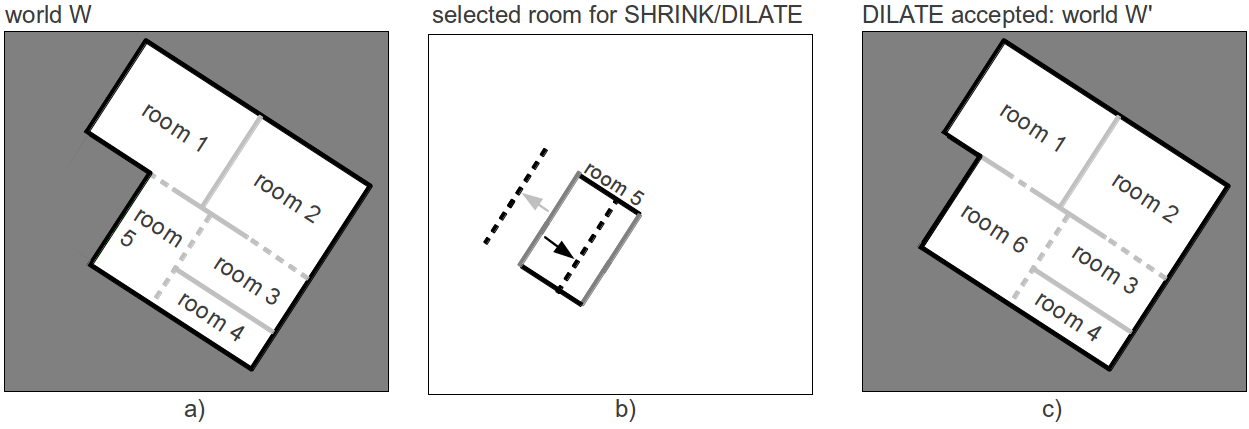}
    \caption{SHRINK/DILATE kernel pair. a) The current world $W$. b) The room 5 is selected. Here, the two gray walls are more likely to be chosen, because they match the map worse than the other two black walls. We assume that the left gray wall is selected for SHRINK/DILATE. The light-gray arrow points to one DILATE possibility, whereas the black arrow points to a SHRINK possibility.  c) After the accepted DILATE action, the world $W^{'}$ is generated.}
   	\label{figure:shrink}
\end{figure*}

Once the room is selected, one wall $w_j$ needs to be drawn from its four walls. Following the same idea, we define a new weight $v_{w_j}$ for sampling the wall:

\begin{equation}
v_{w_j}=\left\{\begin{array}{cl}
 \frac{1}{\omega_{w_j}}, &\frac{1}{\omega_{w_j}} \leq h_v\\
h_v, &\textrm{otherwise},
\end{array}
\right.
\end{equation}
where $\omega_{w_j}$ is the wall weight defined in (\ref{equ:wallweight}), and $h_v$ is a predefined threshold. Again, the wall is drawn according to $v_{w_j}$, as done in (\ref{equ:normalize}) to (\ref{equ:sampling}). After the wall is selected, we try to shift it parallel to its original orientation using a bias that is drawn from a zero-mean Gaussian distribution. In principle, the algebraic sign of the selected bias decides whether a SHRINK or a DILATE is proposed, e.g. if a positive sign proposes a SHRINK, then a negative sign will propose a DILATE. In general, SHRINK and DILATE sub-kernel have both 50\% chance to be proposed. An example of the SHRINK/DILATE kernel pair is shown in Fig.~\ref{figure:shrink}.

\subsubsection{ALLOCATE}

A door detector which is based on connected-components analysis \cite{chang04} proposes door candidates for the sub-kernel ALLOCATE. We draw one door candidate from the set of all candidates according to their weights. Here, the weight $\omega_g$ of a door $g$ is similar to the weight of walls $\omega_{w_j}$ that is defined in (\ref{equ:wallweight}):

\begin{equation}
\omega_{g}=\frac{n^{'}(g)}{l(g)},\label{equ:door weight}
\end{equation}
where $l(g)$ is calculated the same as in (\ref{equ:length}), and $n^{'}(g)$ is computed as follows:

\begin{equation}
n^{'}(g)=\sum\limits_{(x,y)\in g}t^{'}(x,y),
\end{equation}

where

\begin{equation}
t^{'}(x,y)=\left\{\begin{array}{lc}
1,\quad C_{M}(x,y)=2,\\
0,\quad \textrm{otherwise}.
\end{array}
\right.
\end{equation}

Using the weight $\omega_g$, one door candidate is drawn from the set of all detected candidates, as done in (\ref{equ:normalize}) to (\ref{equ:sampling}). Then, the MH algorithm decides whether this door will be accepted. In the following, we detail how to detect door candidates.

According to the indisputable fact that doors must be located on walls in the real world, we search doors along walls in the structured world in the following steps:

\begin{enumerate}
	\item[1)] Expand each wall in its perpendicular direction, so that a rectangular area is created out of each wall. Note that each wall should be shortened before the expansion, so that the extended area does not overlap each other within one room. 
	\item[2)] Detect the overlap of these rectangles using connected-components analysis, because the overlap area indicates on which wall the potential door candidates can be found. Localize the wall part that has caused the overlap.
	\item[3)] Divide the localized wall part averagely into several small segments, so that each segment is equally long. Because each of these segments could be a part of a door, we weight them according to (\ref{equ:door weight}) and try to combine the verified segments to build a door (we define a segment as a verified segment, if its weight is bigger than certain weight threshold).
	\item[4)] Combine the verified segments on each wall, if the distance between them is lower than certain distance threshold. Find the corresponding part of the combined segments on both walls and use them as door candidates.
\end{enumerate}

\begin{figure*}[htb]
	\centering
  	\includegraphics[width=0.8\textwidth]{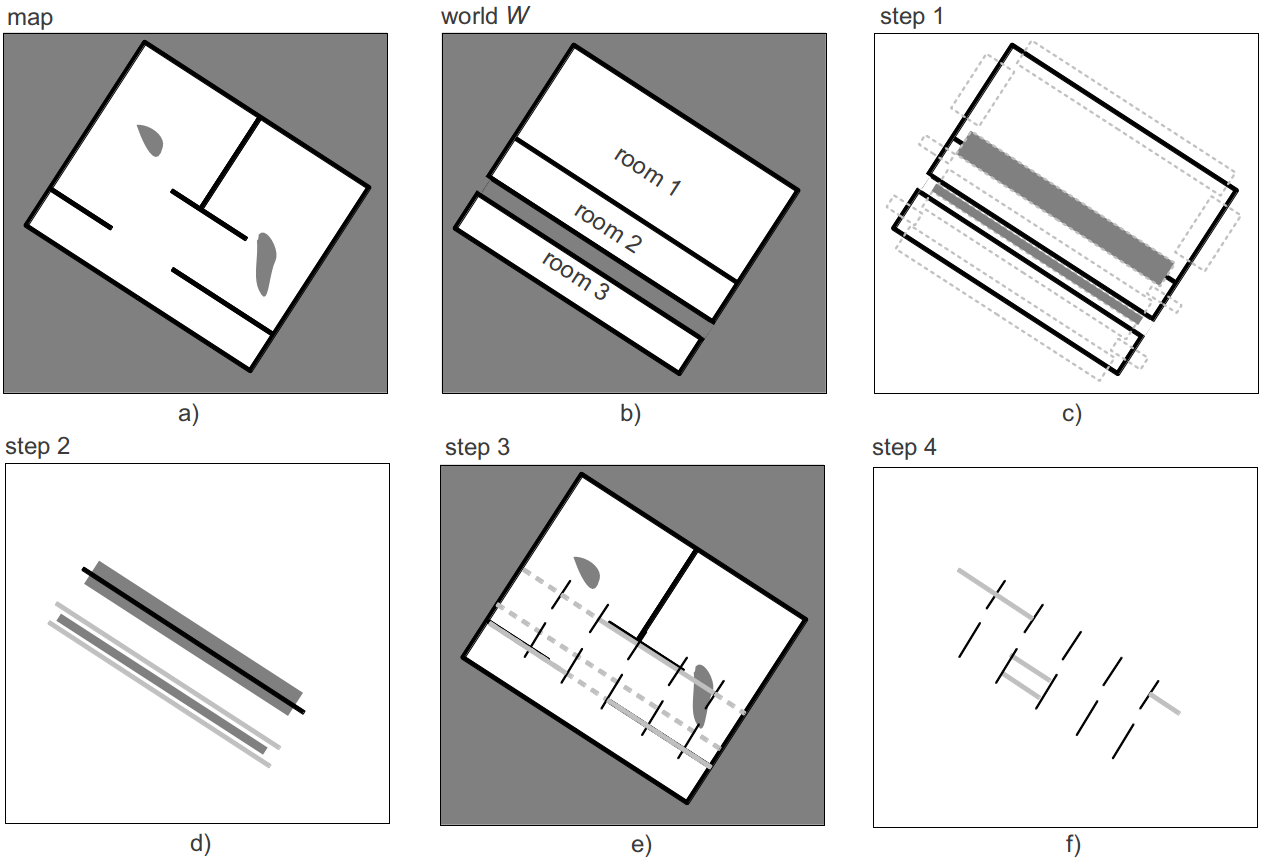}
    \caption{ALLOCATE sub-kernel. a) The occupancy grid map. b) The current world $W$ with no assigned doors. c) Step 1: each wall is shortened and expanded. The expanded areas are marked by light-gray dashed rectangles. The rectangles filled with gray show the overlap areas. d) Step 2: the two pairs of walls that have caused the overlap areas are localized. Note that the black line indicates a pair of overlapping walls, and the other pair of walls is shown in light-gray. e) Step 3: divide the two pairs of walls into six segments, using the black lines. Weight each segment according to how well they match the map. Verified segments are shown in dashed lines, other segments are drawn in solid lines. f) Step 4: combine those verified segments, between which the distance is under certain threshold. Find the corresponding part of the combined segments and use them as door candidates (light-gray lines). Note, all the segments on one wall are detected as verified segment in step 3, which means this whole wall forms a big combined segment, but the final door candidate must correspond to the door candidate on the other wall. That is the reason why only a small door candidate is detected on this wall pair.}
   	\label{figure:allocate}
\end{figure*}

The above process is demonstrated in Fig. \ref{figure:allocate}.

%Since the performance of ALLOCATE strongly depends on the data likelihood of the current world $W$, i.e. the better $W$ matches the map, the better the detected door candidates could be, thus, the sub-kernel ALLOCATE is not activated right at the beginning of the algorithm (see Table \ref{TAB:Transition}). 
After a successful ALLOCATE action, one door is assigned to two rooms, and each assigned door contains the following information: ID of the two rooms, ID of the walls on which the door is located, grid cell indices of the door. 
%Note that one door could have different grid cell index with respect to the two walls, on which this door is detected.

\subsubsection{REMOVE and DELETE}

The sub-kernel REMOVE and DELETE have similar functionality, which is to cancel one existing member room and one assigned door respectively. There are no special discriminative methods used for these two sub-kernels. They just draw one member from the corresponding set (existing rooms or assigned doors) and propose to cancel this member, then the MH algorithm decides whether this proposal is accepted. Following the idea that the worse a member matches the map, the more likely it should be canceled, we use the weight $b_r$ defined in (\ref{equ: new room weight}) for room sampling. Similarly, we define a new weight $z_g$ for door sampling:

\begin{equation}\label{equ: new door weight}
z_g=\left\{\begin{array}{cl}
 \frac{1}{\omega_g}, &\frac{1}{\omega_g} \leq h_g\\
h_g, &\textrm{otherwise},
\end{array}
\right.
\end{equation}
where $\omega_g$ is the door weight defined in (\ref{equ:door weight}), and $h_g$ is a predefined threshold. 

\subsection{Proposal Probability $Q(W^{'}|W)$ and $Q(W|W^{'})$}

The proposal probability $Q(W^{'}|W)$ describes how probable it is that the world $W$ transits to the world $W^{'}$, and by contrast, $Q(W|W^{'})$ is the probability for transiting back to the world $W$, given the world $W^{'}$. Intuitively, $Q(W^{'}|W)$ is the product of the normalized weight of the selected elements (room candidate, splitting line, wall etc.) in the corresponding MC sub-kernel defined in the previous section. For instance, in the ADD or REMOVE sub-kernel, $Q(W^{'}|W)$ is equal to the corresponding normalized weight of the selected room candidate or that of the selected member room. $Q(W^{'}|W)$ in ALLOCATE and DELETE can be calculated similarly to that in ADD and REMOVE respectively. In SPLIT, $Q(W^{'}|W)$ is the product of the corresponding normalized weight of the selected member room and that of the selected splitting line. In SHRINK/DILATE, $Q(W^{'}|W)$ is product of three terms: the corresponding normalized weight of the member room, that of the selected wall and that of the generated Gaussian bias. Similarly, $Q(W^{'}|W)$ of MERGE is calculated as the product of the corresponding normalized weight of the first room and that of the second room.

Compared with $Q(W^{'}|W)$, the calculation of $Q(W|W^{'})$ is less intuitive, because the back transition is virtual and must be defined. For ADD, $Q(W|W^{'})$ should perform the same function as the sub-kernel REMOVE, namely, the world $W'$ transits back to the world $W$ by canceling the room that is added in the transition from $W$ to $W'$, thus $Q(W|W^{'})$ of ADD should be the normalized weight of the added room in the sub-kernel REMOVE. $Q(W|W^{'})$ of REMOVE can also be calculated as the normalized weight of the room, that is canceled in the transition from $W$ to $W'$, in the sub-kernel ADD. Analogously, $Q(W|W^{'})$ of SHRINK, DILATE, MERGE, SPLIT, ALLOCATE and DELETE can be calculated in a style similar to $Q(W^{'}|W)$ in their corresponding reverse sub-kernels. In addition, the SHRINK/DILATE pair just tries to move one wall of the selected room using a relatively small bias, thus the resulting world $W'$ is similar to $W$. For computational simplicity, we assume that $Q(W^{'}|W)$ and $Q(W|W^{'})$ are equal in the SHRINK/DILATE pair.

\begin{table*}[htb]
	\centering
	\begin{tabular}{|c||*{5}{c|}}\hline
	\backslashbox{sub-kernel}{iteration $\beta$}
	%&\makebox[3em]{$\beta\leq$600}&\makebox[3em]{$600<\beta\leq$1000}&\makebox[3em]{$<$2000}&\makebox[3em]{$<$3000}&\makebox[3em]{$<$4000}\\\hline\hline
	
	&\makebox{$\beta\leq$1000}&\makebox{$1000<\beta\leq$4000}&\makebox{$\beta>$4000}\\\hline\hline

	%%&0&1&2\\\hline\hline
	ADD 	&0.8	&0.05	&0.05	\\\hline
	REMOVE	&0.2	&0.05	&0.05	\\\hline
	SPLIT	&0		&0.2 	&0.2	\\\hline
	MERGE	&0		&0.2 	&0.2	\\\hline
	SHRINK 	&0		&0.25	&0.2	\\\hline
	DILATE 	&0		&0.25	&0.2	\\\hline
	ALLOCATE&0		&0 		&0.05	\\\hline
	DELETE	&0		&0 		&0.05 	\\\hline
	\end{tabular}
	\caption{Transition probabilities $\Phi(W'|W)$ of MC sub-kernels.}
	\label{TAB:Transition}
\end{table*}
\section{Experiments}
\label{exp}

We apply our algorithm to several occupancy grid maps. The selection probabilities of the MC sub-kernels are listed in Table~\ref{TAB:Transition}. Here, the selection probabilities depend partially on the iteration index $\beta$. At the beginning ($\beta\leq1000$), the world $W$ does not contain much information about the map, so we mainly apply ADD to propose new rooms into the world, using the WBR  detector. For $\beta>1000$, the selection probability of ADD is set to be very low (0.05), because most part of the map is already explained during the initial exploration ($\beta\leq1000$). In addition, we activate SPLIT, MERGE, SHRINK and DILATE to change the form of the member rooms of the world. For $\beta>4000$, we activate ALLOCATE and DELETE, so that the connectivity information is explored and attached to the world. Moreover, the FSR detector is also activated for $\beta>4000$ to detect left-over free space regions. Table~\ref{TAB:Transition} effectively implements a heuristic scheduling policy.

\begin{figure*}[htb]
\centering
\includegraphics[width=0.45\textwidth]{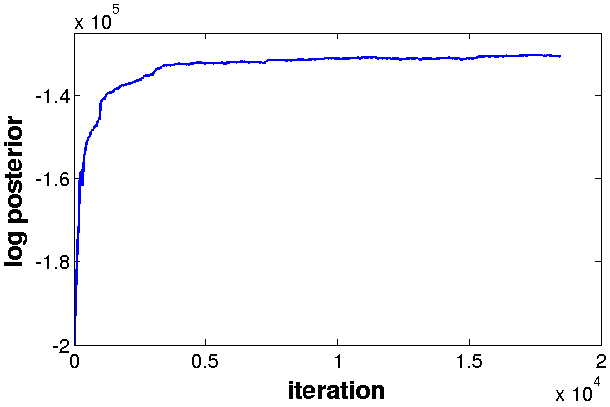}
\includegraphics[width=0.45\textwidth]{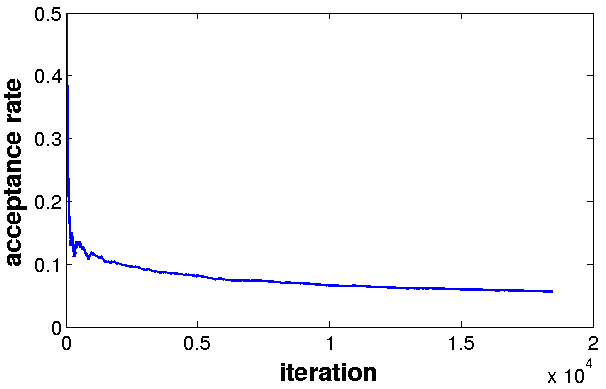}
\caption{The typical development of the posterior probability $P(W|M)$ (left) for the input map shown in Fig.~\ref{figure:bremen}  and the acceptance rate of the proposed state transitions (right) along the Markov chain development in terms of iterations.}
\label{figure:plots1}
\end{figure*}

Fig.~\ref{figure:plots1} depicts the process of Markov chain convergence by showing the evolution of the log posterior $\log(P(W|M))$ along the development of the Markov chain on the left side. In an initial burn-in process the chain quickly approaches its target distribution $P(W|M)$. This is indicated by the rapid increase of the posterior in the beginning. We can also see the discontinuities at the iteration 1000 and 4000 which show the effect of the scheduling policy. The jump at iteration 1000 is a consequence of the activation of new transition kernels that greatly improve the system's capability to structurally adapt the world state $W$ to the observations in the map. After the initial phase, the chain samples from $P(W|M)$ and produces samples that are slight variations of the world $W$ and do not significantly improve the situation any more. 

This convergence process can also be seen by looking at the development of the acceptance rate of the Metropolis-Hastings algorithm (see Fig.~\ref{figure:plots1}, right). In the early phases, the acceptance rate is comparatively high, which means that most of the transitions proposed by $Q(W'|W)$ are accepted, since they correspond to a significantly improved explanation of the map $M$ by the model $W'$. Towards the end, the acceptance rate stabilizes on a low level.

\begin{figure*}[htb]
	\centering
  	\includegraphics[width=0.8\textwidth]{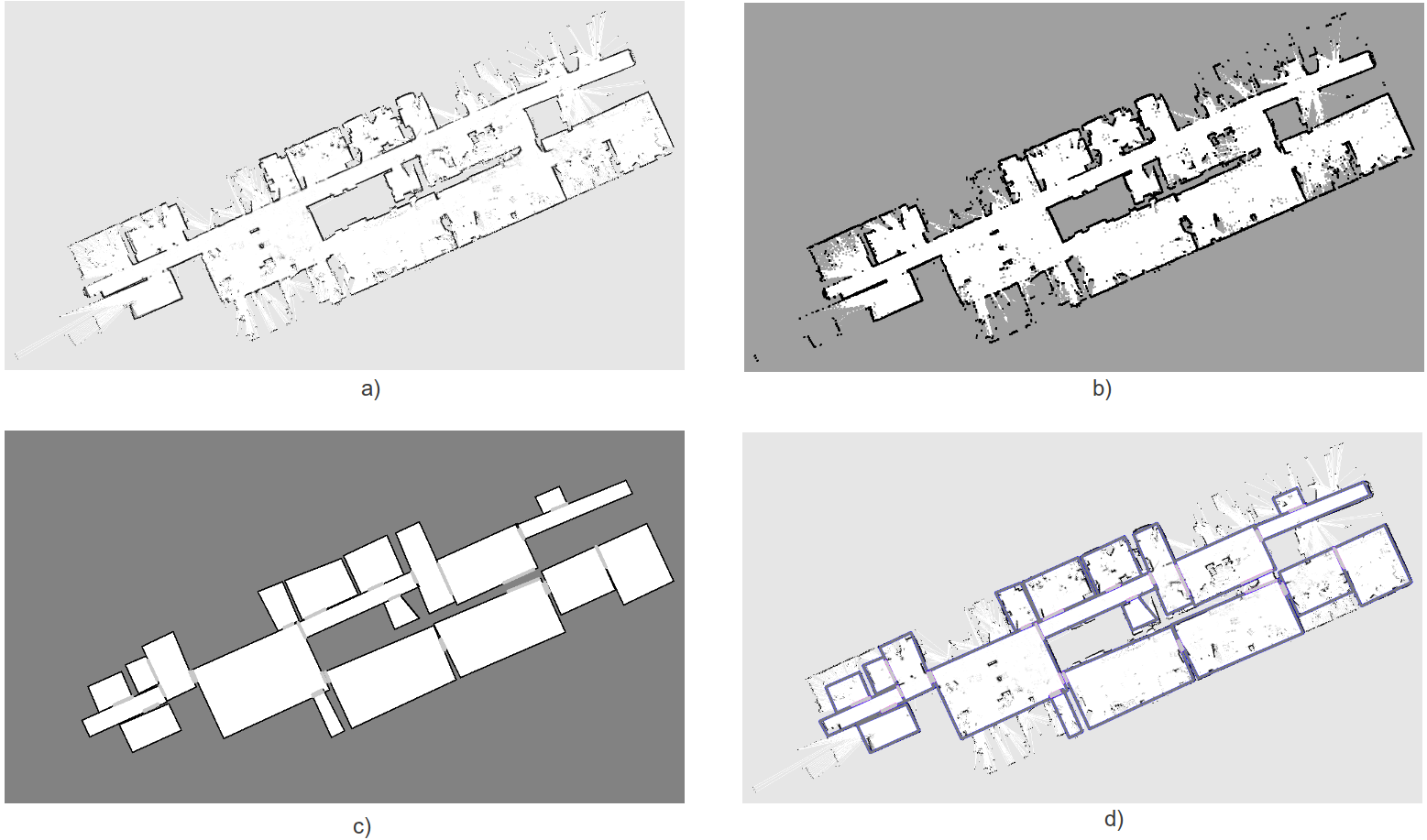}
    \caption{Analysis of the ``ubremen-cartesium" dataset \cite{Radish}. a) The occupancy grid map $M$. b) The classified map $C_M(x,y)$ with three intensity values (black=wall, grey=unexplained, white=free). c) The analyzed world $W$ (black=wall, gray=unknown, white=free, light-gray=door). d) Walls (gray) and doors (light-gray) of the world $W$ drawn into the original map $M$.}
   	\label{figure:bremen}
\end{figure*}

Fig.~\ref{figure:bremen} shows a typical result of the overall process. Here, part a) shows an original input occupancy map $M$ \cite{Radish}. Part b) shows the classified map $C_M(x,y)$ that is defined in (\ref{equ:classify}), with black, gray and white indicating occupied, unexplained and free cells respectively. Part c) visualizes the world state $W$ representing our structured semantic model. Here black, gray, white and light-gray show the wall, unknown, free and door way cells respectively. In part d), walls (gray) and doors (light-gray) of the world $W$ are directly plotted onto the original input map $M$, so as to give a more intuitive comparison.

Since we effectively produce samples from $P(W|M)$ representing the distribution of probable worlds $W$ given the observation $M$, different samples represent different explanations of the data using the modeling structures available for constructing $W$. In our case, these are rectangular rooms and doorways. For demonstration purposes we purposefully chose a map, that is not fully compliant with these assumptions. Therefore various alternative explanations should produce equally good results.

\begin{figure*}[htb]
	\centering
  	\includegraphics[width=0.8\textwidth]{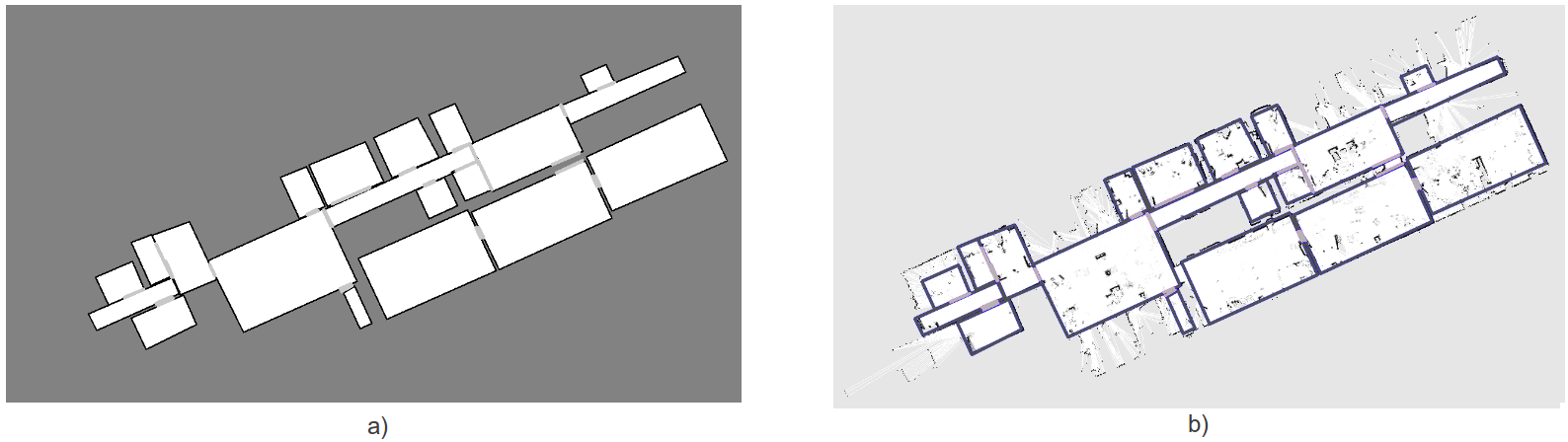}
    \caption{Alternative explanation 1 of the ``ubremen-cartesium" dataset \cite{Radish}.}
   	\label{figure:bremen2}
\end{figure*}

\begin{figure*}[htb]
	\centering
  	\includegraphics[width=0.8\textwidth]{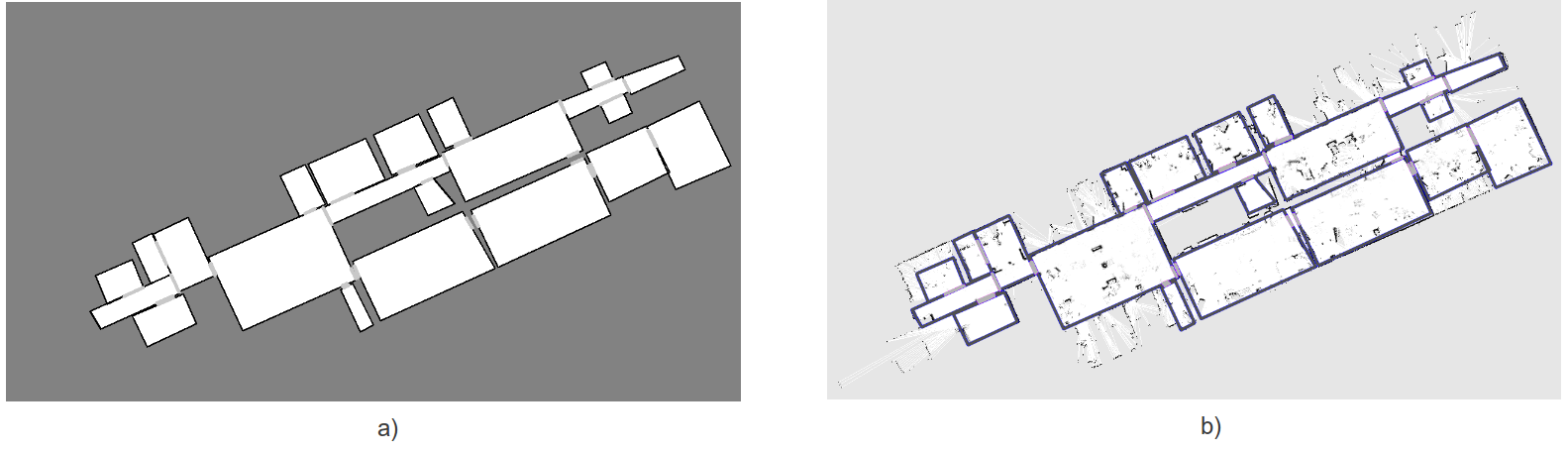}
    \caption{Alternative explanation 2 of the ``ubremen-cartesium" dataset \cite{Radish}.}
   	\label{figure:bremen3}
\end{figure*}

\begin{figure*}[htb]
	\centering
  	\includegraphics[width=0.8\textwidth]{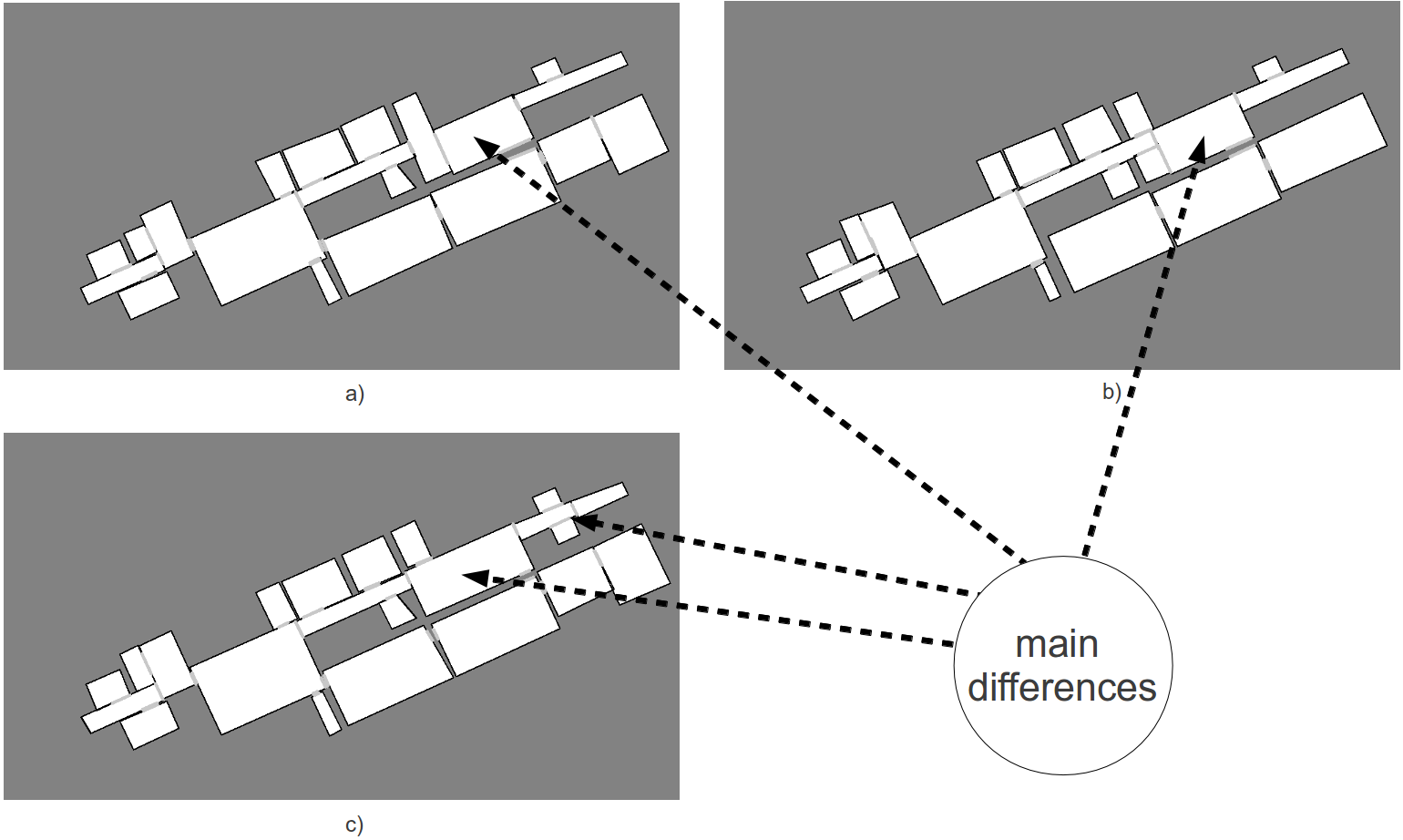}
    \caption{Three explanations of the ``ubremen-cartesium" dataset \cite{Radish}.}
   	\label{figure:bremen4}
\end{figure*}

Fig.~\ref{figure:bremen2} and Fig.~\ref{figure:bremen3} show two high-likelihood samples drawn from $P(W|M)$, i.e.\ two alternative explanations of the input map used in Fig.~\ref{figure:bremen}. Three such samples are compared in Fig.~\ref{figure:bremen4}, where the main differences are pointed out by arrows. A human observer, using a more complex understanding of typical architecture and also of furniture that is currently not modeled, would probably select the model in Fig.~\ref{figure:bremen3}.

\begin{figure*}[htb]
	\centering
  	\includegraphics[width=0.8\textwidth]{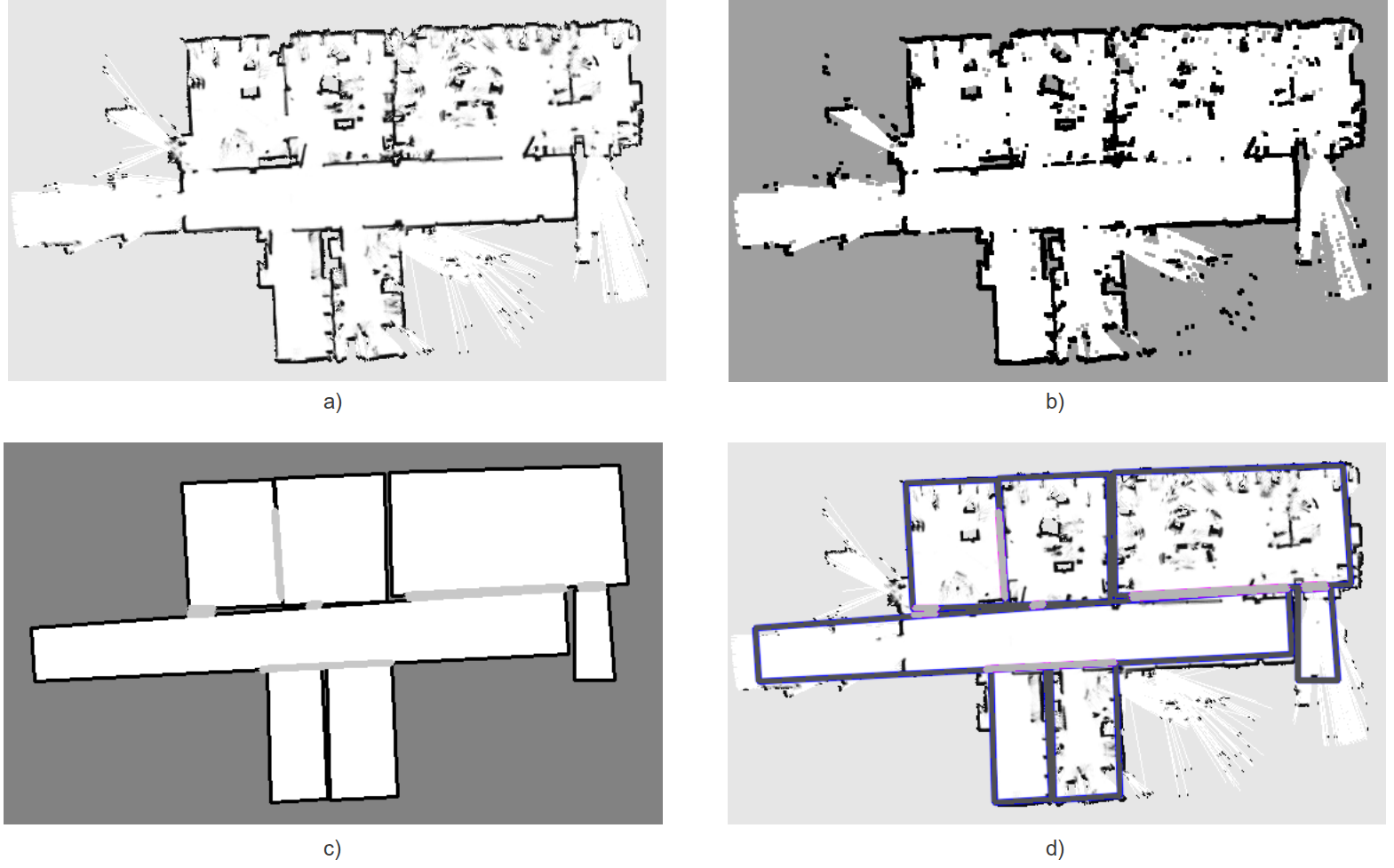}
    \caption{Analysis of the ``albert-b-laser" dataset \cite{Radish}. a) The occupancy grid map. b) The classified map (black=wall, grey=unexplained, white=free). c) The analyzed world (black=wall, gray=unknown, white=free, light-gray=door). d) Walls (gray) and doors (light-gray) of the world drawn into the original map.}
   	\label{figure:map}
\end{figure*}

A result for a different input map \cite{Radish} is illustrated in Fig.~\ref{figure:map}. Compared with the map used in Fig.~\ref{figure:bremen}, this map is relatively simple.

\begin{figure*}[!htb]
\centering
\includegraphics[width=0.45\textwidth]{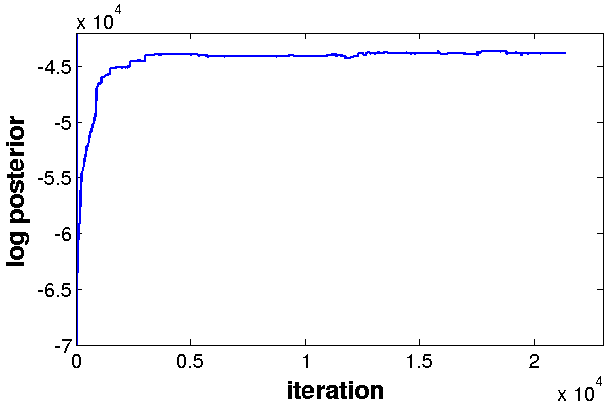}
\includegraphics[width=0.45\textwidth]{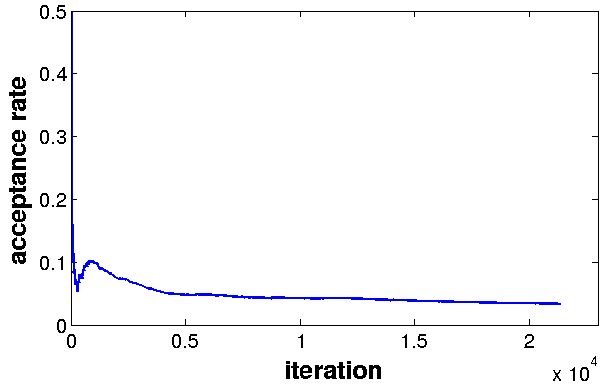}
\caption{The typical development of the posterior probability $P(W|M)$ (left) for the input map shown in Fig.~\ref{figure:gridmap1}  and the acceptance rate of the proposed state transitions (right) along the Markov chain development in terms of iterations.}
\label{figure:plots2}
\end{figure*}

As before Fig.~\ref{figure:plots2} shows the progression of chain convergence in terms of log-posterior and acceptance rates. Again, we can clearly identify the burn-in phase and the effects of our scheduling policy at iterations 1000 and 4000.

The computational cost strongly depends on the size of the occupancy grid map, and the major part of the computation is spent in the evaluation of the generative model. The computation speed of analyzing the map in Fig. \ref{figure:bremen} (size:1237$\times$672) is around 30 iterations per second (ips), and in general it takes about 10000 to 20000 iterations, until the Markov chain reaches a good state, so the computation time on the current PC is around 5 to 10 minutes. By contrast, the map used in Fig. \ref{figure:map} is much smaller (size:556$\times$322), and the same computer reaches around 140 ips, which leads to an overall computation time of 1 to 2 minutes. Currently, we use a single-threaded implementation, where at each iteration only one sub-kernel is tested for the sampling. One of the most important features of the Markov chain is that the current state is only dependent on the previous one, therefore, it is theoretically possible to do multiple tests using different sub-kernels at each iteration, then only the successful test results are saved for the sampling. Using today's powerful off-the-shelf multi-core CPUs, this idea can be easily realized and should lead to a much less computation time.

\section{Summary and Outlook}
\label{sum}

This paper proposes a new approach for automatically extracting semantic information from more or less preprocessed sensor data. We propose to do this by means of a probabilistic generative model and MCMC-based reasoning techniques. Our work differs from previous semantic mapping approaches, that mostly use various classification methods in a bottom-up fashion to label either spatial regions or places based on context or that assign semantic labels directly to portions of the observations. Instead we construct an abstracted semantic and top-down representation of the domain under the consideration: a classical indoor environment consisting of several rooms, that are connected by doorways.

We use Bayesian reasoning to build this semantic map, so that it is aligned with the preprocessed sensor observations, that a robot made during an environment exploration and mapping stage. This introduces a bottom-up path into the approach and employs data driven discriminative environment feature detectors to analyze the continuous noisy sensor observations. The semantic environment model that we generate, is structured similarly to a scene graph and is perfectly suited for any higher level reasoning and communication purposes. 
 
Currently, we assume rooms with a rectangular shape, which however does not imply that our approach is restricted to this room type only. The general idea behind this is to demonstrate the exploitation of abstract (uncertain) rules on how the environment might be structured. Adhering to these rules helps the robot to interpret the noisy sensor data more correctly. In fact, other room types can be introduced to the approach in that we update the prior, add discriminative methods for proposing rooms of other types and implement functionalities for carrying out new geometrical operations (e.g. for SHRINK/DILATE and SPLIT/MERGE). However, the general approach will be the same. 

While we currently generate representations that more or less resemble a classical floor plan (including semantics however), the extension of our work to more functionally enhanced representations (e.g. differentiating several room types based on the context, adding other types of concepts like general objects or furniture) will be pursued in the future. It is also straight forward to extend the concept towards 3D environment representations. A second line of research will address the integration of this type of semantic knowledge into the perception procedures at the run time of the robot.

\section*{Acknowledgements}

This work is accomplished with the support of the Technische Universit\"at M\"unchen - Institute for Advanced Study, funded by the German Excellence Initiative.

We thank the anonymous reviewers for their valuable comments that were greatly helpful for improving the quality of the paper.

The ``ubremen-cartesium" dataset and the ``albert-b-laser" dataset used in the experiments were obtained from the Robotics Data Set Repository (Radish)~\cite{Radish}. We thank Cyrill Stachniss for providing these two datasets.



%% References

\section*{References}
\bibliographystyle{elsarticle-num}
\bibliography{semslam_elsevier2012}



%\begin{thebibliography}{00}
%
%%% \bibitem must have the following form:
%%%   \bibitem{key}...
%%%
%
%\bibitem{z1} \url{http://cres.usc.edu/radishrepository/view-all.php}
%
%\end{thebibliography}

\break

\vspace{1cm}
\begin{wrapfigure}{l}{1.5cm}
\includegraphics[width=2cm]{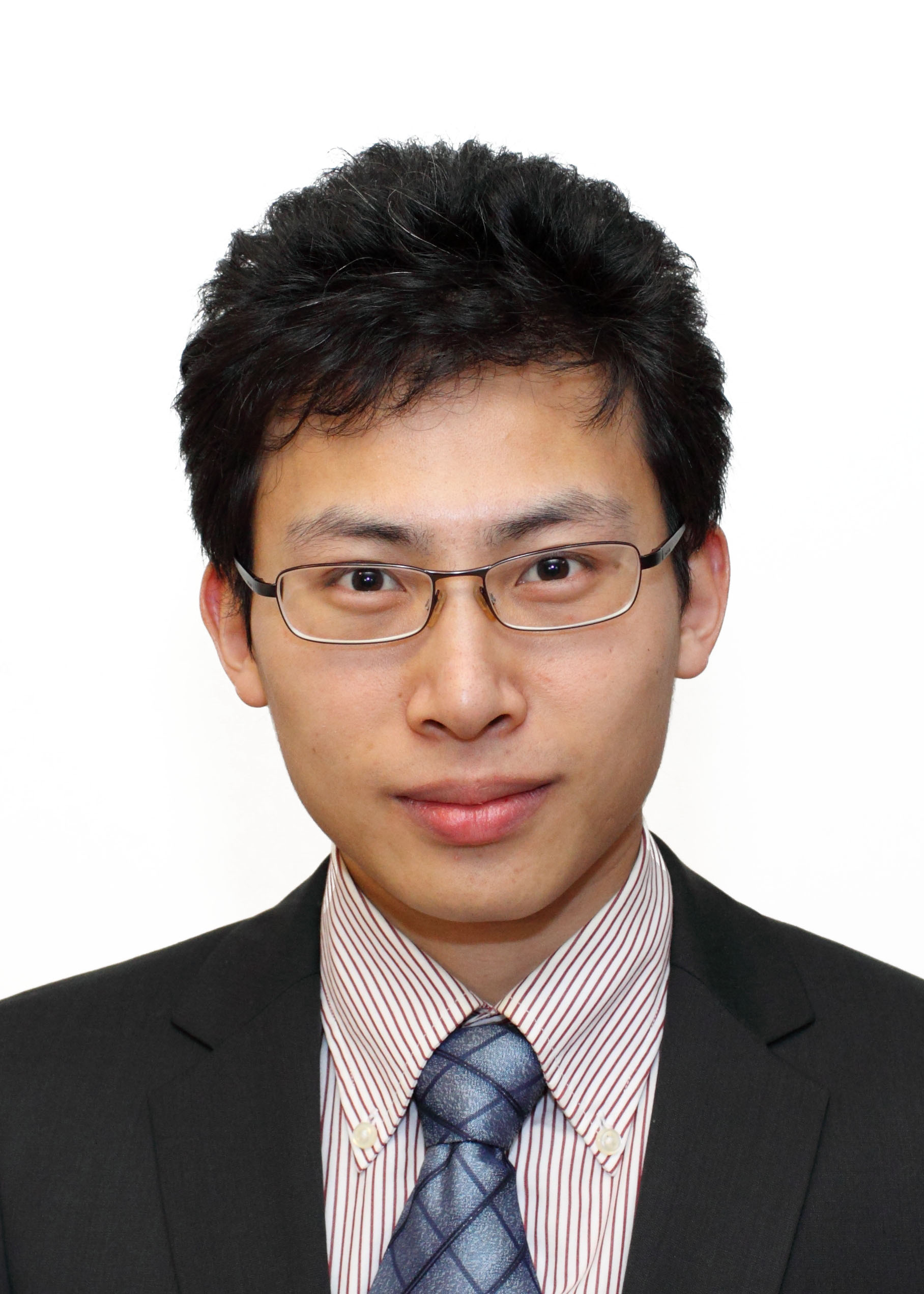}
\end{wrapfigure}
\small \noindent\textbf{Ziyuan Liu} received his B.E. degree in Mechatronics from the TongJi University, Shanghai, China, in 2008. He received his M.S. degree in 2010 from the Institute of Automatic Control Engineering at Technische Universit\"at M\"unchen, Munich, Germany. Currently, he is a Ph.D. candidate at the Institute of Automatic Control Engineering at Technische Universit\"at M\"unchen. His research interests are semantic perception and sampling based inference methods.\\

\begin{wrapfigure}{l}{1.5cm}
\includegraphics[width=2cm]{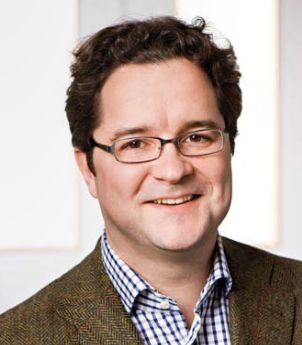}
\end{wrapfigure}

\small \noindent\textbf{Georg von Wichert} received his Diploma (MSc) in Electrical and Control Engineering from Darmstadt University of Technology in 1992. From 1992 to 1998 he was a research and teaching assistant at the Institute of Control Engineering at Darmstadt University of Technology. In Darmstadt he also received the Ph.D. degree in Electrical Engineering in 1998. Since 1998 his is with Siemens Corporate Research and Technologies, where he currently holds the position of the program manager for Cognitive Autonomous System. At the same time he is a fellow of the Institute for Advanced Study at Technische Universit\"at M\"unchen.

\end{document}